\pdfoutput=1

\documentclass[11pt]{article}

\usepackage[preprint]{acl}

\usepackage{times}
\usepackage{latexsym}

\usepackage[T1]{fontenc}

\usepackage[utf8]{inputenc}

\usepackage{microtype}

\usepackage{graphicx}

\usepackage{enumitem} 

\usepackage{xcolor}
\usepackage{mdframed}

\usepackage[colorinlistoftodos,prependcaption]{todonotes}

\title{Proofread: Fixes All Errors with One Tap}

\author{Renjie Liu\thanks{Equal contribution, alphabetical order. Correspondence to \texttt{\{renjieliu,zhangyx,yunzhu\}@google.com}.}, Yanxiang Zhang$^*$, Yun Zhu$^*$, Haicheng Sun, Yuanbo Zhang, \\
        \textbf{Michael Xuelin Huang, Shanqing Cai, Lei Meng, Shumin Zhai} \\
        Google Inc.}

\begin{document}
\maketitle

\begin{abstract}

The impressive capabilities in Large Language Models (LLMs) provide a powerful approach to reimagine users' typing experience. This paper demonstrates Proofread, a novel Gboard feature powered by a server-side LLM in Gboard, enabling seamless sentence-level and paragraph-level corrections with a single tap. We describe the complete system in this paper, from data generation, metrics design to model tuning and deployment. To obtain models with sufficient quality, we implement a careful data synthetic pipeline tailored to online use cases, design multifaceted metrics, employ a two-stage tuning approach to acquire the dedicated LLM for the feature: the Supervised Fine Tuning (SFT) for foundational quality, followed by the Reinforcement Learning (RL) tuning approach for targeted refinement. Specifically, we find sequential tuning on Rewrite and proofread tasks yields the best quality in SFT stage, and propose global and direct rewards in the RL tuning stage to seek further improvement. Extensive experiments on a human-labeled golden set showed our tuned PaLM2-XS model achieved 85.56\% good ratio. 
We launched the feature to Pixel 8 devices by serving the model on TPU v5 in Google Cloud, with thousands of daily active users.  Serving latency was significantly reduced by quantization, bucket inference, text segmentation, and speculative decoding. Our demo could be seen in \href{https://youtu.be/4ZdcuiwFU7I}{Youtube}.


\end{abstract}

\section{Introduction}

Gboard is an statistical-decoding-based keyboard on mobile devices developed by Google. Decoding \cite{ouyang2017mobile} is necessary due to the error-prone process of "fat finger" touch input on small screens. According to \citet{azenkot2012touch}, the per-letter error rate is around 8\%-9\% without decoding. 

Gboard provides various error correction features, some active (automatic) and other passive (require the user's further manual action and selection) to provide a smooth typing experience \cite{ouyang2017mobile}. Active key correction (KC), and active auto correction (AC), word completions and next-word predictions support the user to type the current word and next word by fixing user typos and providing multiple word candidates in the suggestion bar or inline (smart compose). Post correction (PC) supports fixing errors in last one or more committed words. Furthermore, The more passive Spell Checker and Grammar Checker supported by small-sized logistic regression and seq2seq models respectively detect the possible errors in committed sentences and mark them with red underlines, users can fix the errors by clicking the incorrect words and commit the correct words from the displayed candidates.

There are two types of user experience limitations with the existing correction approaches. First, users still have to type relatively slowly and accurately to avoid making too many or too severe errors that the small (but instantly fast) on-device correction models such as KC,  AC and PC cannot handle due to their limited ability to model longer-span context. 
Second, users need to manually engage in the multi-step passive correction features, such as the grammar checker and the spell checker,  to correct the committed words one after another.

Supervising the committed words while typing and fixing errors sequentially by editing after commit take users' cognitive and visual-motor resources and slow down their typing speed. One desired pattern of fast typing users of Gboard is to focus on keyboard only without checking the committed words while typing. To this end, a high quality sentence-or-higher-level correction feature is often called for, in order to help those ''fast and sloppy'' users who prefer to focus on typing then switch to error corrections at a higher level. 



\begin{figure*}
    \centering
    \includegraphics[width=15cm]{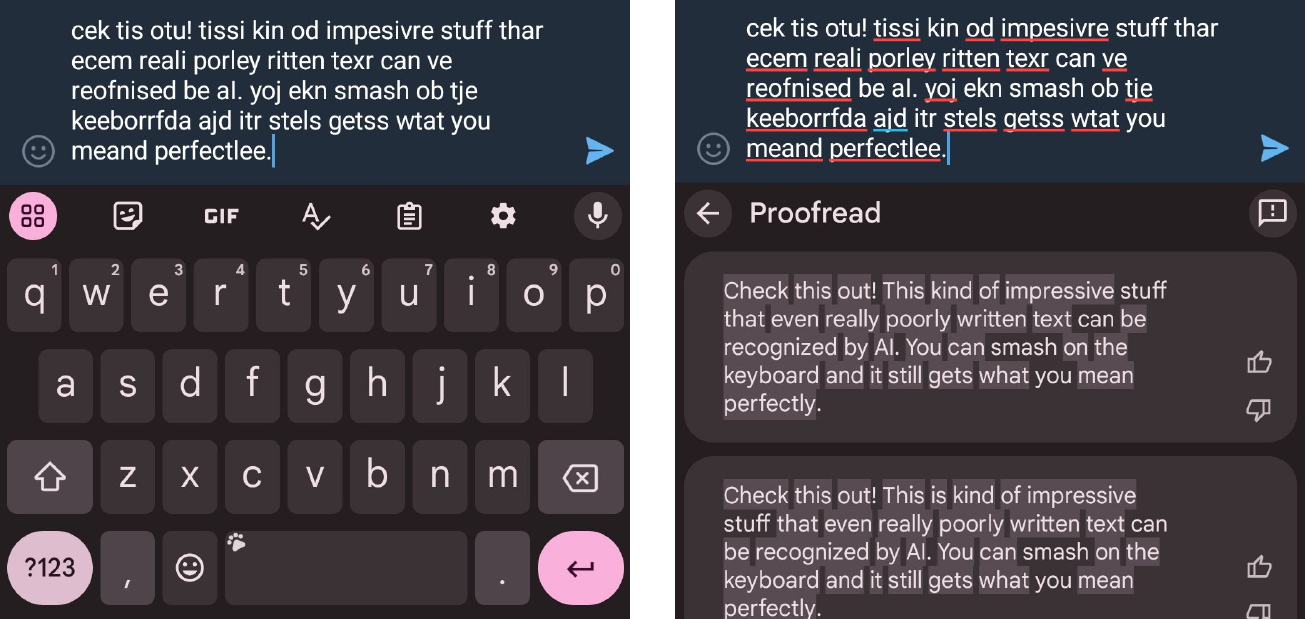}
    \caption{Proofread demo on a heavy corrupted text, the feature is triggered by clicking the "A" button in the left figure.}
    \label{fig:proofread_demo}
\end{figure*}


In this paper, we propose the \textbf{Proofread} feature to alleviate the pain points of fast typers by providing the sentence-level and paragraph-level error fixes with only one-tap. Proofread falls into the area of Grammatical Error Correction (GEC), which has a long history of research from rule-based to statistical approaches to neural network models \cite{bryant2023grammatical}. The astonishing capability growth of Large Language Models (LLMs), offers a new opportunity to unlock the high quality sentence-level grammar fixes.

We present the entire system to tune and serve the LLM model behind Proofread in this paper. The system consist of four parts, data generation, metrics design, model tuning and model serving. Firstly, dataset is generated by a carefully designed error synthetic framework which integrates errors frequently made on keyboard to simulate the users' input, several further steps are conducted to ensure the data distribution is close to Gboard domain 
maximally. Secondly, several metrics are designed to measure the model from various dimensions. As the answers are always not unique specifically for long examples, the metric combined with grammar error existence check and same meaning check based on LLMs are considered as the key metrics for comparing the model quality. Thirdly, inspired by InstructGPT \cite{ouyang2022training}, Supervised Fine-tuning followed by the Reinforcement Learning (RL) tuning is adopted to obtain the LLM dedicated for Proofread feature. 
Results suggested that our rewrite task tuning and reinforcement learning recipe significantly improves the proofreading performance of the foundation models. 
To reduce the serving cost, we build our feature on top of the medium sized LLM PaLM2-XS, which could be fit int a single TPU v5 after 8-bit quantization. We further optimize latency with  bucket keys, segmentation and speculative decoding \cite{leviathan2023fast}.
Our model now is launched to benefit thousands of users with Pixel 8 devices.

Figure \ref{fig:proofread_demo} exhibits our model quality on one extreme corrupted case from Andrej Karpathy\footnote{https://twitter.com/karpathy/status/1725553780878647482}, which indicates our tuned model is strong enough to handle various of heavy typo errors made by users.

The contribution of this paper can be summarized as follows:

\begin{itemize}[nosep]
    \item We propose the \textbf{Proofread} feature supported by the high quality LLM to boost the user typing experiences of Gboard. We finally launched the feature to real users with Pixel 8 devices, thousands of users benefit from it daily.
    \item We design and implement the whole system from data generation, metrics design to model tuning and deployment.
    \item We obtain a high quality model with cautiously synthetic data generation, multiple phased supervised fine-tuning and RL tuning. Specifically, we propose the Global Reward and Direct Reward in RL tuning stage, which improve the model significantly. Results shows that RL tuning could help reduce the grammar error significantly and thus the Bad ratio of PaLM2-XS model is reduced by 5.74\% relatively.
    \item We deploy the model to TPU v5 in Cloud with highly optimized latency acquired by quantization, buckets, input segmentation and speculative decoding. Our results suggested that speculative decoding reduced the median latency by 39.4\%. 
\end{itemize}

\section{Related Work}

\subsection{Controllable Text Generation}

Controllable text generation using transformer-based pre-trained language models has become a rapid growing yet challenging new research hotspot \cite{zhang2023survey}. Proofread falls into this scope with the requirement of modifying the input to fix the grammar errors without changing the original intention in the corrupted text.

Lots of applications could inherit from controllable text generation. \citet{shu2023rewritelm, zhu2023towards} focus on text rewrite tasks, including paraphrasing \cite{xu2012paraphrasing, siddique2020unsupervised}, style transfer \cite{riley2020textsettr, zhang2020parallel, reif2021recipe} and sentence fusion \cite{mallinson2022edit5} and so on. 
Similarly, Text editing \cite{malmi2022text} task also covers a wide range of sub-tasks such as paraphrasing, style transfer, spelling and grammatical error correction \cite{napoles2017jfleg}, formalization \cite{rao2018dear}, simplification \cite{xu2016optimizing} and elaboration \cite{logan2021fruit}.

Unlike these mentioned works, our paper only addresses a single application -- Proofread but provides systematic approaches that optimize the model from different perspective such as quality, latency and resource usage.  

\subsection{Grammatical Error Correction (GEC)}

Proofread falls into the area of GEC. \citet{bryant2023grammatical} offers a comprehensive survey of the history and the current state of GEC. Specifically, before LLM, the popular solutions of GEC are edit-based approaches which corrections are applied on a sequence labelling \cite{omelianchuk2020gector} or sequence-to-sequence basis \cite{stahlberg2020seq2edits}.

The recent studies to apply LLM to GEC mainly focus on prompting the LLM rather than supervised fine-tuning. \citet{wu2023chatgpt} compares ChatGPT to Grammarly, \citet{coyne2023analyzing} compares GPT-3.5 and GPT-4 to two GEC system on English benchmarks. \citet{davis2024prompting} conducts a more comprehensive study by evaluating seven open-source and three commercial LLMs on four established GEC benchmarks.

Following the LLM trend, our system is built upon latest LLM backbone. But we apply instruction tuning approach to customize the LLM. 

\subsection{Instruction Tuning(IT)}

Instruction tuning has been proven to be an efficient approach to boost model performance and generalization to unseen tasks \cite{chung2022scaling, sanh2021multitask}. Reinforcement learning with human feedback (RLHF) is leveraged to further extend instruction tuning in InstructGPT \cite{ouyang2022training}. Reinforcement learning with AI feedback (RLAIF) \cite{bai2022constitutional} could alleviate the heavy human preference data dependency, \citet{zhu2023towards, cheng2021heuristic} further replace the reward model in RLAIF with a heuristic model, which will be adopted in this paper to boost the quality. 
Our instruction tuning approach is inspired by the previous works and also follows the 2-step tuning process. 
We designed the synthetic data generation and RL strategy in a heuristic way that favors the proofreading task. 

\subsection{Latency Optimization}

Numerous techniques aim to speed up inference of LLMs, which can be categorized into two major lines according to the focus point. The first line mainly focuses on model or algorithm side, including model compression with pruning \cite{xia2023sheared} and sparsity \cite{xia2023flash}, quantization \cite{dettmers2208llm}, small model design \cite{timiryasov2023baby, liu2024mobilellm}, attention computation optimizations like low-rank approximation \cite{katharopoulos2020transformers}, sparse attention \cite{roy2021efficient} etc. 

The other line explores acceleration along with hardwares, exemplified works includes FlashAttention \cite{dao2022flashattention}, FlexGen \cite{sheng2023flexgen}, which considers hardware scheduling and weight movement, and speculative decoding \cite{leviathan2023fast, chen2023accelerating}, which leverages the parallism of hardwares to pre-conduct computation with sampling method.

We adopt quantization and speculative decoding to accelerate the inference speed in the model deployment.

\section{Dataset}

\begin{figure*}
    \centering
    \includegraphics[width=15cm]{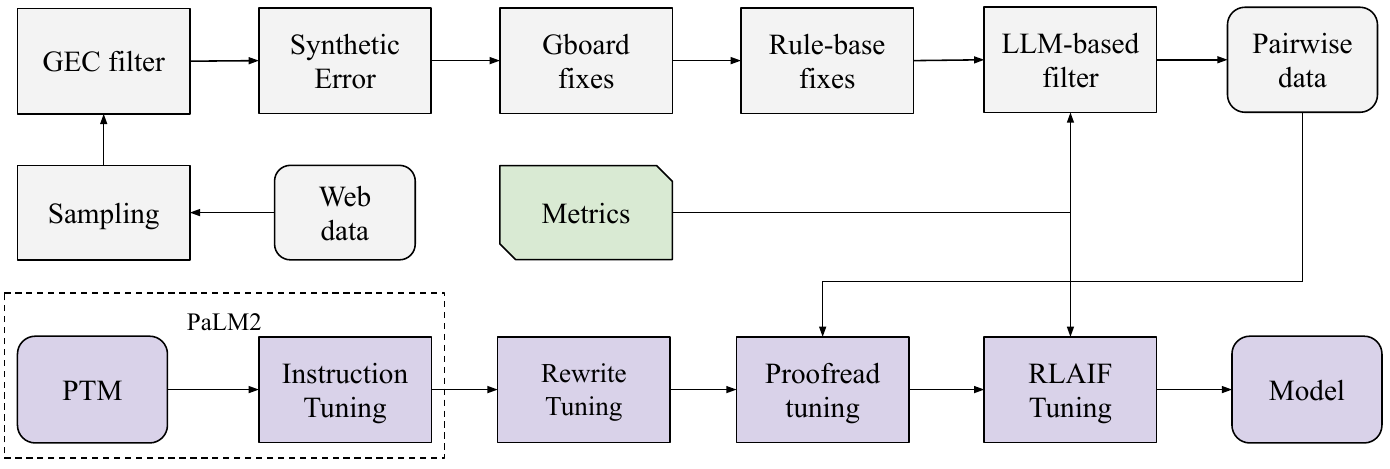}
    \caption{Data synthesis and Model tuning pipeline}
    \label{fig:tuning_phases}
\end{figure*}

The upper half of Figure \ref{fig:tuning_phases} illustrates the pipeline to generate the dataset. We initially sample data from the web crawled dataset, which is then processed by a GEC model to fix the grammar errors. Each item in the dataset consists of a source sentence with several possible reference sentences.

Grammar errors are then synthesized into the source sentence to simulate users' inputs, various kinds of errors which frequently happen in Gboard real scenarios are involved in this step, including: 



\begin{itemize}[nosep]
    \item character omission, e.g., "hello" as "hllo"
    \item character insertion, e.g., "hello" as "hpello"
    \item transposition, e.g., "hello" as "hlelo"
    \item double tap, e.g., "hello" as "heello"
    \item omit double characters, e.g., "hello" as "helo"
    \item Gaussian-based positional errors, e.g., "hello" as "jello"
\end{itemize}

To align the dataset with real use cases, the data with synthetic errors are then passed to the Gboard simulator to fix errors by leveraging Gboard's built-in literal decoding, KC and AC functions. Moreover, several heuristic rules were then applied to fix cases such as emoji/emoticons alignment, date time formatting, and URL patterns. 


The last step is to filter the noise data by utilizing LLM with careful designed instructions to avoid polluting the model. Data is diagnosed by various dimensions, including:


\begin{itemize}[nosep]
    \item The reference sentence still has errors remaining.
    \item The reference sentence itself is not fluent or clear enough.
    \item The reference sentence has different meaning as the source sentence.
    \item The reference sentence has different tones, aspects and tense from the source sentence.
\end{itemize}

To maximally benefit the model quality, the criteria above coordinates to the metrics defined in the following section.

An example of the synthetic dataset is showcased below:

\begin{mdframed}[backgroundcolor=white]
\textbf{Source}: "Good Moning! hey si, how. a u dou?"\\
\textbf{Reference1}: "Good morning! Hey sir, how are you doing?"\\
\textbf{Reference2}: "Good morning! Hey sister, how are you doing?"
\end{mdframed}

Moreover, part of the examples labeled by human rater are used as the golden set for evaluation.

\section{Metrics}

It's of key importance to define the correct metrics which are aligned to user experiences online before the feature goes to public. In this section, several metrics are designed to measure the model quality.

Given the three elements, input (corrupted text), answer(predicted candidate from the model) and target(ground truth), we present the following metrics.

\begin{itemize}[nosep]
    \item \textbf{EM} / Exact Match Ratio: ratio of answer equal to target exactly.
    \item \textbf{NEM} / Normalized Exact Match Ratio : ratio of answer equal to target ignoring capitalization and punctuation.
    \item \textbf{Error} Ratio: ratio of answer containing grammar errors, which is conducted by LLM with specific instruction.
    \item \textbf{Diff} Meaning Ratio: ratio of answer and target don't have the same meaning, which is also conducted by LLM with specific instruction.
    \item \textbf{Good} Ratio: ratio of answer without grammar error and has the same meaning with target.
    \item \textbf{Bad} Ratio: ratio of answer either have grammar error or has different meaning with target.
\end{itemize}

From the definition, the Good/Bad ratios combining Error check and Diff Meaning check, are the primary metrics due to their robustness from LLM. The bad ratio is a bit more important as it portrays how much the users could tolerate the errors made by model.
The combination of Error / Diff Meaning checks is also leveraged as the reward in RL phase of model tuning. EM/NEM ratios are referenced as supporting indicators as they are too strict for examples with multiple references.

\section{Model tuning}

The lower half of Figure \ref{fig:tuning_phases} illustrates the tuning steps of the model for Proofread. We start from instruction-tuned models. PaLM2-XS model from \citet{anil2023palm} is the candidate model.


\subsection{Supervised Fine-tuning}

The initial step after choosing the checkpoint is to fine-tune the model on the rewrite dataset, which contains hundreds of text rewriting tasks from \citet{shu2023rewritelm, zhu2023towards}. We assume that fine-tuning on similar tasks is beneficial to the final quality of Proofread. After that, the models are fine-tuned on synthetic dataset.


The evaluation results of multiple phased tuning are displayed in Table \ref{tab:eval_results}. It's natural to observe that after supervised fine-tuning on the synthetic dataset, the model quality can be largely improved from 65.48\% to 83.80\%. An interesting finding is that though fine-tuning on Rewrite dataset degrades the quality, sequential fine-tuning on Rewrite and Proofread datasets yields the best results with Good ratio 84.68\% and Bad ratio 15.32\%, which we argue the robustness is enhanced by the large size of the combined dataset, by comparing M2 and M3, we can further conclude that Rewrite tuning contributes to the intent preservation.



\begin{table*}[ht]
    \setlength{\tabcolsep}{2pt}
    \centering
    \caption{The metrics of PaLM2-XS tuned on various phases on the Golden dataset. The upper half focuses on the supervised fine-tuning, and model variants in Reinforcement Learning phase are listed in the lower half.}
    \label{tab:eval_results}
    \begin{tabular}{cccccccc}
        \hline
        Model ID & PaLM variant & EM(\%) & NEM(\%) & Good(\%) & Bad(\%) & DIFF(\%) & ERROR(\%)   \\
        \hline
        M0& PALM2-XS & 29.96 & 45.80 & 65.48 & 34.52 & 18.56 & 30.32  \\
        M1& M0 + Rewrite & 23.44 & 40.90 & 59.48 & 40.52 & 19.04 & 37.04  \\
        M2& M0 + Proofread & 37.88 & 55.30 & 83.80 & 16.20 & 12.08 & 8.12 \\
        M3& M0 + Rewrite + Proofread & 39.16 & 56.20 & 84.68 & 15.32 & 10.60 & 9.68  \\
        \hline
        M4& M3 + RL Global Reward & 35.92 & 53.80 & 85.24 & 14.76 & 11.12 & 6.80 \\
        M5& M3 + RL Direct Reward & 32.20 & 50.20 & \textbf{85.56} & 14.44 & 11.52 & 5.68 \\
        M6& M5 + Large weight on KL & 39.08 & 55.40 & 84.76 & 15.24 & 10.96 & 8.88 \\
        \hline
    \end{tabular}
\end{table*}

\subsection{Reinforcement Learning}

RLAIF is leveraged with heuristic rewards in our model tuning following \citet{zhu2023towards} to avoid relying on human labelers. Two alternative heuristic rewards based on LLM are designed in this paper.

\begin{itemize}[nosep]
    \item \textbf{Global Reward}: With few-shot examples, the LLM tells whether a candidate is a good fix of the corrupted inputs. 
    \item \textbf{Direct Reward}: As the goal is to improve the Good Ratio, we directly convert the grammar error check and diff meaning check into rewards, both relying on LLM and will be combined as the final reward. This requires the ground truth included in the example.
\end{itemize}

Proximal Policy Optimization (PPO) \cite{schulman2017proximal} is facilitated to optimize the model. KL divergence is involved to help model keep the ability to recover the original text \cite{peters2010relative, mitchell2023emulator}.

The second part of Table \ref{tab:eval_results} exhibits the results of RL tuning with different rewards. It's observed that the Bad ratio of PaLm2-XS model could be improved by 3.65\% and 5.74\% relatively through applying the RL with Global Reward and RL with Direct Reward respectively. 


Specifically, RL excels at reducing the grammatical error but struggles to maintain meaning alignment between prediction and ground truth by comparing M3, M4 and M5. We argue that the optimizing meaning is inherently more subjective and complex comparing than grammar. Additionally, RL reduces the EM and NEM ratios, indicating a shift in the output distribution for both correct and incorrect cases. While increasing the KL divergence penalty can mitigate this (See M5 and M6), it doesn't significantly improve the Good/Bad ratios. We suspect the defined metrics might have inherent conflicts. Future work will reply on online metrics and real user data to drive further improvement.


\section{Model Serving}

Google's TPUv5e \cite{googletpuv5e} is utilized to serve the Proofread model, which is the latest Google TPU chip with 16GB HBM. 8-bit quantization is facilitated to reduce the memory footprint and latency without observing quality degradation.


In the context of our research, which predominantly focuses on deployment within chat applications, it has been observed that the average sentence length seldom exceeds 20 words. Consequently, we have established a discrete set of bucket keys, specifically [16, 32, 64, 128], to categorize the input data accordingly. Furthermore, we have calibrated the temperature parameter to a value of 0.3, aiming to maintain a constrained level of creativity in the proofreading outcomes, thereby ensuring relevance and coherence.

To be capable of handling more extensive documents, a systematic approach is employed wherein the document is segmented into individual paragraphs. All paragraphs are then processed in parallel, allowing for a more manageable and efficient analysis.

\begin{figure}
    \centering
    \includegraphics[width=7.5cm]{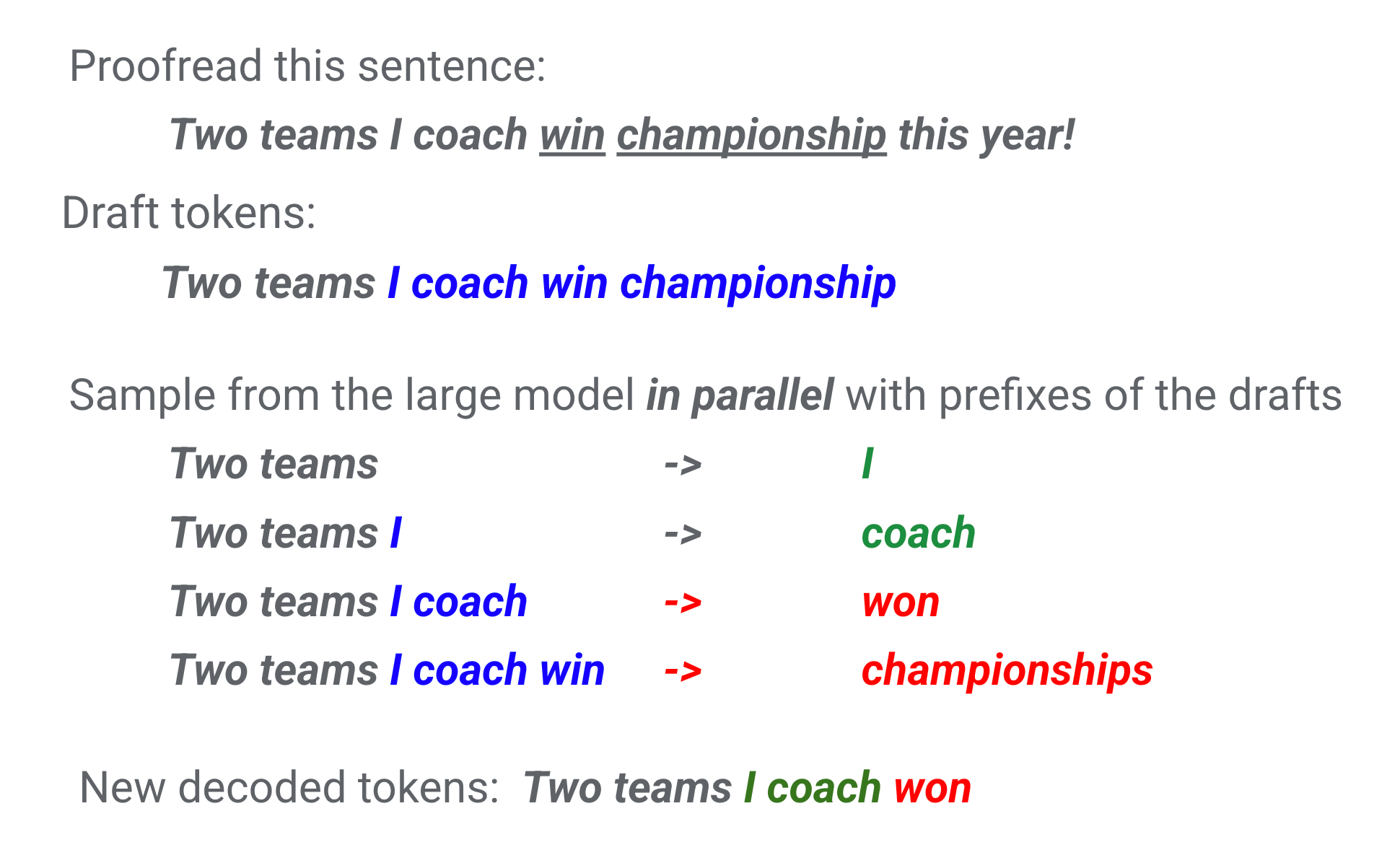}
    \caption{Example of the speculative decoding process for the proofreading task. The words in green color are selected draft by the LLM. This process speeds up the decoding without quality regression.}
    \label{fig:spec}
\end{figure}

Additionally, our methodology incorporates the use of speculative decoding \cite{leviathan2023fast}, complemented by heuristic drafter models that are tailored to align with user history patterns. 
Under our proofreading case, the initial input would naturally fit into the speculative draft so external drafter models are needed.  We share an example to illustrate the process in Figure~\ref{fig:spec}.
This innovative approach significantly contributes to the reduction of operational costs. Through empirical evaluation in Table~\ref{tab:spec}, we have recorded a 39.4\% reduction on median latency per serving request, as measured on Tensor Processing Unit (TPU) cycles, underscoring the efficiency of our system in real-time applications.

\begin{table}
    \centering
    \caption{Latency improvement with speculative decoding.}
    \label{tab:spec}
    \begin{tabular}{cc}
        \hline
         Decoding & Latency (ms) \\
        \hline
        Baseline &  314.4 \\
        + speculative  & 190.6 (-39.4\%) \\
        \hline
    \end{tabular}
\end{table}

\section{Conclusions}

This paper presents a novel Proofread feature implemented within Gboard, powered by a carefully refined LLM. Our work demonstrates the significant potential of LLMs to enhance the users' typing experiences by providing high-quality sentence- and paragraph-level corrections. We detailed our comprehensive approach, encompassing synthetic data generation pipeline aligned with real-world use cases, multifaceted metrics design, two-stage model tuning (multiple phased SFT followed by RL) and the efficient model deployment.

Specifically, our findings reveal that rewrite task tuning benefited the SFT model by enhancing the meaning alignment ability of the model. Additionally, we discovered the value of global and direct rewards during RL tuning, which could further improve the model by reduce grammar errors significantly. Rigorous experiments demonstrated that our tuned PaLM2-XS model achieved an impressive 85.56\% good ratio and 14.44\% bad ratio. The successful deployment of the model on TPU v5, leveraging optimizations such as quantization, bucket inference and speculative decoding, highlights its real-world viability.

This work underscores the transformative power of LLMs in the realm of user input experiences. Future research directions include leveraging real-user data, multilingual adaption, personalized assistance for diverse writing styles and privacy-preserving on-device solutions. This technology has the potential to fundamentally improve how we interact with our devices.

\section*{Acknowledgments}

The authors would like to thank Ananda Theertha Suresh, Jae Ro, Ziteng Sun, Shankar Kumar, Lan Wei, Yuki Zhong, Zhe Su, JinDong Chen for their insightful discussions and support.

\bibliography{kb_proofread_demo}

\begin{thebibliography}{43}
\expandafter\ifx\csname natexlab\endcsname\relax\def\natexlab#1{#1}\fi

\bibitem[{Anil et~al.(2023)Anil, Dai, Firat, Johnson, Lepikhin, Passos,
  Shakeri, Taropa, Bailey, Chen et~al.}]{anil2023palm}
Rohan Anil, Andrew~M Dai, Orhan Firat, Melvin Johnson, Dmitry Lepikhin,
  Alexandre Passos, Siamak Shakeri, Emanuel Taropa, Paige Bailey, Zhifeng Chen,
  et~al. 2023.
\newblock Palm 2 technical report.
\newblock \emph{arXiv preprint arXiv:2305.10403}.

\bibitem[{Azenkot and Zhai(2012)}]{azenkot2012touch}
Shiri Azenkot and Shumin Zhai. 2012.
\newblock Touch behavior with different postures on soft smartphone keyboards.
\newblock In \emph{Proceedings of the 14th international conference on
  Human-computer interaction with mobile devices and services}, pages 251--260.

\bibitem[{Bai et~al.(2022)Bai, Kadavath, Kundu, Askell, Kernion, Jones, Chen,
  Goldie, Mirhoseini, McKinnon et~al.}]{bai2022constitutional}
Yuntao Bai, Saurav Kadavath, Sandipan Kundu, Amanda Askell, Jackson Kernion,
  Andy Jones, Anna Chen, Anna Goldie, Azalia Mirhoseini, Cameron McKinnon,
  et~al. 2022.
\newblock Constitutional ai: Harmlessness from ai feedback.
\newblock \emph{arXiv preprint arXiv:2212.08073}.

\bibitem[{Bryant et~al.(2023)Bryant, Yuan, Qorib, Cao, Ng, and
  Briscoe}]{bryant2023grammatical}
Christopher Bryant, Zheng Yuan, Muhammad~Reza Qorib, Hannan Cao, Hwee~Tou Ng,
  and Ted Briscoe. 2023.
\newblock Grammatical error correction: A survey of the state of the art.
\newblock \emph{Computational Linguistics}, 49(3):643--701.

\bibitem[{Chen et~al.(2023)Chen, Borgeaud, Irving, Lespiau, Sifre, and
  Jumper}]{chen2023accelerating}
Charlie Chen, Sebastian Borgeaud, Geoffrey Irving, Jean-Baptiste Lespiau,
  Laurent Sifre, and John Jumper. 2023.
\newblock Accelerating large language model decoding with speculative sampling.
\newblock \emph{arXiv preprint arXiv:2302.01318}.

\bibitem[{Cheng et~al.(2021)Cheng, Kolobov, and
  Swaminathan}]{cheng2021heuristic}
Ching-An Cheng, Andrey Kolobov, and Adith Swaminathan. 2021.
\newblock Heuristic-guided reinforcement learning.
\newblock \emph{Advances in Neural Information Processing Systems},
  34:13550--13563.

\bibitem[{Chung et~al.(2022)Chung, Hou, Longpre, Zoph, Tay, Fedus, Li, Wang,
  Dehghani, Brahma et~al.}]{chung2022scaling}
Hyung~Won Chung, Le~Hou, Shayne Longpre, Barret Zoph, Yi~Tay, William Fedus,
  Yunxuan Li, Xuezhi Wang, Mostafa Dehghani, Siddhartha Brahma, et~al. 2022.
\newblock Scaling instruction-finetuned language models.
\newblock \emph{arXiv preprint arXiv:2210.11416}.

\bibitem[{Coyne et~al.(2023)Coyne, Sakaguchi, Galvan-Sosa, Zock, and
  Inui}]{coyne2023analyzing}
Steven Coyne, Keisuke Sakaguchi, Diana Galvan-Sosa, Michael Zock, and Kentaro
  Inui. 2023.
\newblock Analyzing the performance of gpt-3.5 and gpt-4 in grammatical error
  correction.
\newblock \emph{arXiv preprint arXiv:2303.14342}.

\bibitem[{Dao et~al.(2022)Dao, Fu, Ermon, Rudra, and
  R{\'e}}]{dao2022flashattention}
Tri Dao, Dan Fu, Stefano Ermon, Atri Rudra, and Christopher R{\'e}. 2022.
\newblock Flashattention: Fast and memory-efficient exact attention with
  io-awareness.
\newblock \emph{Advances in Neural Information Processing Systems},
  35:16344--16359.

\bibitem[{Davis et~al.(2024)Davis, Caines, Andersen, Taslimipoor,
  Yannakoudakis, Yuan, Bryant, Rei, and Buttery}]{davis2024prompting}
Christopher Davis, Andrew Caines, {\O}istein Andersen, Shiva Taslimipoor, Helen
  Yannakoudakis, Zheng Yuan, Christopher Bryant, Marek Rei, and Paula Buttery.
  2024.
\newblock Prompting open-source and commercial language models for grammatical
  error correction of english learner text.
\newblock \emph{arXiv preprint arXiv:2401.07702}.

\bibitem[{Dettmers et~al.(2022)Dettmers, Lewis, Belkada, and
  Zettlemoyer}]{dettmers2208llm}
Tim Dettmers, Mike Lewis, Younes Belkada, and Luke Zettlemoyer. 2022.
\newblock Gpt3. int8 (): 8-bit matrix multiplication for transformers at scale.
\newblock \emph{Advances in Neural Information Processing Systems},
  35:30318--30332.

\bibitem[{Google(2023)}]{googletpuv5e}
Google. 2023.
\newblock Tpu system architecture.
\newblock
  \url{https://cloud.google.com/tpu/docs/system-architecture-tpu-vm#tpu_v5e}.
\newblock Accessed: 2024-03-15.

\bibitem[{Katharopoulos et~al.(2020)Katharopoulos, Vyas, Pappas, and
  Fleuret}]{katharopoulos2020transformers}
Angelos Katharopoulos, Apoorv Vyas, Nikolaos Pappas, and Fran{\c{c}}ois
  Fleuret. 2020.
\newblock Transformers are rnns: Fast autoregressive transformers with linear
  attention.
\newblock In \emph{International conference on machine learning}, pages
  5156--5165. PMLR.

\bibitem[{Leviathan et~al.(2023)Leviathan, Kalman, and
  Matias}]{leviathan2023fast}
Yaniv Leviathan, Matan Kalman, and Yossi Matias. 2023.
\newblock \href {http://arxiv.org/abs/2211.17192} {Fast inference from
  transformers via speculative decoding}.

\bibitem[{Liu et~al.(2024)Liu, Zhao, Iandola, Lai, Tian, Fedorov, Xiong, Chang,
  Shi, Krishnamoorthi et~al.}]{liu2024mobilellm}
Zechun Liu, Changsheng Zhao, Forrest Iandola, Chen Lai, Yuandong Tian, Igor
  Fedorov, Yunyang Xiong, Ernie Chang, Yangyang Shi, Raghuraman Krishnamoorthi,
  et~al. 2024.
\newblock Mobilellm: Optimizing sub-billion parameter language models for
  on-device use cases.
\newblock \emph{arXiv preprint arXiv:2402.14905}.

\bibitem[{Logan~IV et~al.(2021)Logan~IV, Passos, Singh, and
  Chang}]{logan2021fruit}
Robert~L Logan~IV, Alexandre Passos, Sameer Singh, and Ming-Wei Chang. 2021.
\newblock Fruit: Faithfully reflecting updated information in text.
\newblock \emph{arXiv preprint arXiv:2112.08634}.

\bibitem[{Mallinson et~al.(2022)Mallinson, Adamek, Malmi, and
  Severyn}]{mallinson2022edit5}
Jonathan Mallinson, Jakub Adamek, Eric Malmi, and Aliaksei Severyn. 2022.
\newblock Edit5: Semi-autoregressive text-editing with t5 warm-start.
\newblock \emph{arXiv preprint arXiv:2205.12209}.

\bibitem[{Malmi et~al.(2022)Malmi, Dong, Mallinson, Chuklin, Adamek, Mirylenka,
  Stahlberg, Krause, Kumar, and Severyn}]{malmi2022text}
Eric Malmi, Yue Dong, Jonathan Mallinson, Aleksandr Chuklin, Jakub Adamek,
  Daniil Mirylenka, Felix Stahlberg, Sebastian Krause, Shankar Kumar, and
  Aliaksei Severyn. 2022.
\newblock Text generation with text-editing models.
\newblock \emph{arXiv preprint arXiv:2206.07043}.

\bibitem[{Mitchell et~al.(2023)Mitchell, Rafailov, Sharma, Finn, and
  Manning}]{mitchell2023emulator}
Eric Mitchell, Rafael Rafailov, Archit Sharma, Chelsea Finn, and Christopher~D
  Manning. 2023.
\newblock An emulator for fine-tuning large language models using small
  language models.
\newblock \emph{arXiv preprint arXiv:2310.12962}.

\bibitem[{Napoles et~al.(2017)Napoles, Sakaguchi, and
  Tetreault}]{napoles2017jfleg}
Courtney Napoles, Keisuke Sakaguchi, and Joel Tetreault. 2017.
\newblock Jfleg: A fluency corpus and benchmark for grammatical error
  correction.
\newblock \emph{arXiv preprint arXiv:1702.04066}.

\bibitem[{Omelianchuk et~al.(2020)Omelianchuk, Atrasevych, Chernodub, and
  Skurzhanskyi}]{omelianchuk2020gector}
Kostiantyn Omelianchuk, Vitaliy Atrasevych, Artem Chernodub, and Oleksandr
  Skurzhanskyi. 2020.
\newblock Gector--grammatical error correction: tag, not rewrite.
\newblock \emph{arXiv preprint arXiv:2005.12592}.

\bibitem[{Ouyang et~al.(2022)Ouyang, Wu, Jiang, Almeida, Wainwright, Mishkin,
  Zhang, Agarwal, Slama, Ray et~al.}]{ouyang2022training}
Long Ouyang, Jeffrey Wu, Xu~Jiang, Diogo Almeida, Carroll Wainwright, Pamela
  Mishkin, Chong Zhang, Sandhini Agarwal, Katarina Slama, Alex Ray, et~al.
  2022.
\newblock Training language models to follow instructions with human feedback.
\newblock \emph{Advances in neural information processing systems},
  35:27730--27744.

\bibitem[{Ouyang et~al.(2017)Ouyang, Rybach, Beaufays, and
  Riley}]{ouyang2017mobile}
Tom Ouyang, David Rybach, Fran{\c{c}}oise Beaufays, and Michael Riley. 2017.
\newblock Mobile keyboard input decoding with finite-state transducers.
\newblock \emph{arXiv preprint arXiv:1704.03987}.

\bibitem[{Peters et~al.(2010)Peters, Mulling, and Altun}]{peters2010relative}
Jan Peters, Katharina Mulling, and Yasemin Altun. 2010.
\newblock Relative entropy policy search.
\newblock In \emph{Proceedings of the AAAI Conference on Artificial
  Intelligence}, volume~24, pages 1607--1612.

\bibitem[{Rao and Tetreault(2018)}]{rao2018dear}
Sudha Rao and Joel Tetreault. 2018.
\newblock Dear sir or madam, may i introduce the gyafc dataset: Corpus,
  benchmarks and metrics for formality style transfer.
\newblock \emph{arXiv preprint arXiv:1803.06535}.

\bibitem[{Reif et~al.(2021)Reif, Ippolito, Yuan, Coenen, Callison-Burch, and
  Wei}]{reif2021recipe}
Emily Reif, Daphne Ippolito, Ann Yuan, Andy Coenen, Chris Callison-Burch, and
  Jason Wei. 2021.
\newblock A recipe for arbitrary text style transfer with large language
  models.
\newblock \emph{arXiv preprint arXiv:2109.03910}.

\bibitem[{Riley et~al.(2020)Riley, Constant, Guo, Kumar, Uthus, and
  Parekh}]{riley2020textsettr}
Parker Riley, Noah Constant, Mandy Guo, Girish Kumar, David Uthus, and Zarana
  Parekh. 2020.
\newblock Textsettr: Few-shot text style extraction and tunable targeted
  restyling.
\newblock \emph{arXiv preprint arXiv:2010.03802}.

\bibitem[{Roy et~al.(2021)Roy, Saffar, Vaswani, and
  Grangier}]{roy2021efficient}
Aurko Roy, Mohammad Saffar, Ashish Vaswani, and David Grangier. 2021.
\newblock Efficient content-based sparse attention with routing transformers.
\newblock \emph{Transactions of the Association for Computational Linguistics},
  9:53--68.

\bibitem[{Sanh et~al.(2021)Sanh, Webson, Raffel, Bach, Sutawika, Alyafeai,
  Chaffin, Stiegler, Scao, Raja et~al.}]{sanh2021multitask}
Victor Sanh, Albert Webson, Colin Raffel, Stephen~H Bach, Lintang Sutawika,
  Zaid Alyafeai, Antoine Chaffin, Arnaud Stiegler, Teven~Le Scao, Arun Raja,
  et~al. 2021.
\newblock Multitask prompted training enables zero-shot task generalization.
\newblock \emph{arXiv preprint arXiv:2110.08207}.

\bibitem[{Schulman et~al.(2017)Schulman, Wolski, Dhariwal, Radford, and
  Klimov}]{schulman2017proximal}
John Schulman, Filip Wolski, Prafulla Dhariwal, Alec Radford, and Oleg Klimov.
  2017.
\newblock Proximal policy optimization algorithms.
\newblock \emph{arXiv preprint arXiv:1707.06347}.

\bibitem[{Sheng et~al.(2023)Sheng, Zheng, Yuan, Li, Ryabinin, Chen, Liang,
  R{\'e}, Stoica, and Zhang}]{sheng2023flexgen}
Ying Sheng, Lianmin Zheng, Binhang Yuan, Zhuohan Li, Max Ryabinin, Beidi Chen,
  Percy Liang, Christopher R{\'e}, Ion Stoica, and Ce~Zhang. 2023.
\newblock Flexgen: High-throughput generative inference of large language
  models with a single gpu.
\newblock In \emph{International Conference on Machine Learning}, pages
  31094--31116. PMLR.

\bibitem[{Shu et~al.(2023)Shu, Luo, Hoskere, Zhu, Liu, Tong, Chen, and
  Meng}]{shu2023rewritelm}
Lei Shu, Liangchen Luo, Jayakumar Hoskere, Yun Zhu, Canoee Liu, Simon Tong,
  Jindong Chen, and Lei Meng. 2023.
\newblock Rewritelm: An instruction-tuned large language model for text
  rewriting.
\newblock \emph{arXiv preprint arXiv:2305.15685}.

\bibitem[{Siddique et~al.(2020)Siddique, Oymak, and
  Hristidis}]{siddique2020unsupervised}
AB~Siddique, Samet Oymak, and Vagelis Hristidis. 2020.
\newblock Unsupervised paraphrasing via deep reinforcement learning.
\newblock In \emph{Proceedings of the 26th ACM SIGKDD international conference
  on knowledge discovery \& data mining}, pages 1800--1809.

\bibitem[{Stahlberg and Kumar(2020)}]{stahlberg2020seq2edits}
Felix Stahlberg and Shankar Kumar. 2020.
\newblock Seq2edits: Sequence transduction using span-level edit operations.
\newblock \emph{arXiv preprint arXiv:2009.11136}.

\bibitem[{Timiryasov and Tastet(2023)}]{timiryasov2023baby}
Inar Timiryasov and Jean-Loup Tastet. 2023.
\newblock Baby llama: knowledge distillation from an ensemble of teachers
  trained on a small dataset with no performance penalty.
\newblock \emph{arXiv preprint arXiv:2308.02019}.

\bibitem[{Wu et~al.(2023)Wu, Wang, Wan, Jiao, and Lyu}]{wu2023chatgpt}
Haoran Wu, Wenxuan Wang, Yuxuan Wan, Wenxiang Jiao, and Michael Lyu. 2023.
\newblock Chatgpt or grammarly? evaluating chatgpt on grammatical error
  correction benchmark.
\newblock \emph{arXiv preprint arXiv:2303.13648}.

\bibitem[{Xia et~al.(2023{\natexlab{a}})Xia, Zheng, Li, Zhuang, Zhou, Qiu, Li,
  Lin, and Song}]{xia2023flash}
Haojun Xia, Zhen Zheng, Yuchao Li, Donglin Zhuang, Zhongzhu Zhou, Xiafei Qiu,
  Yong Li, Wei Lin, and Shuaiwen~Leon Song. 2023{\natexlab{a}}.
\newblock Flash-llm: Enabling cost-effective and highly-efficient large
  generative model inference with unstructured sparsity.
\newblock \emph{arXiv preprint arXiv:2309.10285}.

\bibitem[{Xia et~al.(2023{\natexlab{b}})Xia, Gao, Zeng, and
  Chen}]{xia2023sheared}
Mengzhou Xia, Tianyu Gao, Zhiyuan Zeng, and Danqi Chen. 2023{\natexlab{b}}.
\newblock Sheared llama: Accelerating language model pre-training via
  structured pruning.
\newblock \emph{arXiv preprint arXiv:2310.06694}.

\bibitem[{Xu et~al.(2016)Xu, Napoles, Pavlick, Chen, and
  Callison-Burch}]{xu2016optimizing}
Wei Xu, Courtney Napoles, Ellie Pavlick, Quanze Chen, and Chris Callison-Burch.
  2016.
\newblock Optimizing statistical machine translation for text simplification.
\newblock \emph{Transactions of the Association for Computational Linguistics},
  4:401--415.

\bibitem[{Xu et~al.(2012)Xu, Ritter, Dolan, Grishman, and
  Cherry}]{xu2012paraphrasing}
Wei Xu, Alan Ritter, William~B Dolan, Ralph Grishman, and Colin Cherry. 2012.
\newblock Paraphrasing for style.
\newblock In \emph{Proceedings of COLING 2012}, pages 2899--2914.

\bibitem[{Zhang et~al.(2023)Zhang, Song, Li, Zhou, and Song}]{zhang2023survey}
Hanqing Zhang, Haolin Song, Shaoyu Li, Ming Zhou, and Dawei Song. 2023.
\newblock A survey of controllable text generation using transformer-based
  pre-trained language models.
\newblock \emph{ACM Computing Surveys}, 56(3):1--37.

\bibitem[{Zhang et~al.(2020)Zhang, Ge, and Sun}]{zhang2020parallel}
Yi~Zhang, Tao Ge, and Xu~Sun. 2020.
\newblock Parallel data augmentation for formality style transfer.
\newblock \emph{arXiv preprint arXiv:2005.07522}.

\bibitem[{Zhu et~al.(2023)Zhu, Liu, Stahlberg, Kumar, Chen, Luo, Shu, Liu,
  Chen, and Meng}]{zhu2023towards}
Yun Zhu, Yinxiao Liu, Felix Stahlberg, Shankar Kumar, Yu-hui Chen, Liangchen
  Luo, Lei Shu, Renjie Liu, Jindong Chen, and Lei Meng. 2023.
\newblock Towards an on-device agent for text rewriting.
\newblock \emph{arXiv preprint arXiv:2308.11807}.

\end{thebibliography}




\end{document}